\documentclass[3p,times,authoryear]{article} % For LaTeX2e
\usepackage{iclr2021_conference,times}
\usepackage{makecell, multirow}
\usepackage{graphicx}
\usepackage[tableposition=below]{caption}
\usepackage{longtable}
\usepackage{amsmath,amsfonts,bm}

\def\eqref#1{equation~\ref{#1}}
\def\1{\bm{1}}

\DeclareMathAlphabet{\mathsfit}{\encodingdefault}{\sfdefault}{m}{sl}
\SetMathAlphabet{\mathsfit}{bold}{\encodingdefault}{\sfdefault}{bx}{n}

\usepackage{enumitem}
\usepackage{algorithm}
\usepackage{algpseudocode}

\usepackage{hyperref}
\hypersetup{colorlinks,allcolors=black}

\usepackage{fancyhdr}
\usepackage{array}
\usepackage{url}
\usepackage{tabularx}
\usepackage{adjustbox}
\usepackage{array}
\usepackage{float}
\setcitestyle{round,authoryear,comma}
\usepackage{geometry}

\usepackage{array,ragged2e}
\newcolumntype{M}[1]{>{\Centering\arraybackslash}m{#1}}

\makeatletter
\renewcommand\paragraph{\@startsection{paragraph}{4}{0ex}%
   {1mm}%
   {1mm}%
   {\normalfont\normalsize\bfseries}}
\newcommand\mymaketitle{%
  \begin{titlepage}
    \null\vfil\vskip 40\p@
    \begin{center}
      {\@title}
      \vskip 2.5em
      {\large \lineskip .75em \@author \par}
      \vskip 1.5em
      {\large \@date \par}
    \end{center}\par
    \@thanks
    \vfil\null
  \end{titlepage}
}
\makeatother

\usepackage{fancyhdr}

\title{A Hybrid Swarm Intelligence Approach for Optimizing Multimodal Large Language Models Deployment in Edge-Cloud-based Federated Learning Environments}
\author{Gaith Rjoub$^4$, Hanae Elmekki$^1$, Saidul Islam$^1$, Jamal Bentahar$^{2,1,*}$, Rachida Dssouli$^1$ \\
$^1$Concordia Institute for Information Systems Engineering, 
Concordia University, Montreal, Canada \\
$^2$Department of Computer Science,
Khalifa University, Abu Dhabi, UAE\\
$^3$Cheriton School of Computer Science, University of Waterloo, Waterloo, Canada\\
$^4$Faculty of Information Technology, Aqaba University of Technology, Aqaba, Jordan\\
\\\\
\textbf{Contributing Authors' Emails:}\\ jamal.bentahar@concordia.ca;\\ grjoub@aut.edu.jo;\\ saidul.islam@concordia.ca;\\ rachida.dssouli@concordia.ca\\\\
\textbf{$^*$Corresponding Author's Email:}  jamal.bentahar@concordia.ca\\\\
The authors contributed equally to this work.
}
\usepackage{fancyhdr}
\renewcommand{\headrulewidth}{0pt}
\renewcommand{\footrulewidth}{0pt}
\fancypagestyle{first}{%
\fancyhf{} % clear all header and footer fields

\setlength{\headsep}{0.5in}
\renewcommand{\headrulewidth}{0pt}
\renewcommand{\footrulewidth}{0pt}}
\begin{document}

  \begin{center}
\maketitle
\thispagestyle{first}
\end{center}
\begin{abstract}

The combination of Federated Learning (FL), Multimodal Large Language Models (\textit{MLLMs}), and edge-cloud computing enables distributed and real-time data processing while preserving privacy across edge devices and cloud infrastructure. However, the deployment of \textit{MLLMs} in FL environments with resource-constrained edge devices presents significant challenges, including resource management, communication overhead, and non-IID data. 
To address these challenges, we propose a novel hybrid framework wherein \textit{MLLMs} are deployed on edge devices equipped with sufficient resources and battery life, while the majority of training occurs in the cloud. To identify suitable edge devices for deployment, we employ Particle Swarm Optimization (\textit{PSO}), and Ant Colony Optimization (\textit{ACO}) is utilized to optimize the transmission of model updates between edge and cloud nodes.
This proposed swarm intelligence-based framework aims to enhance the efficiency of \textit{MLLM} training by conducting extensive training in the cloud and fine-tuning at the edge, thereby reducing energy consumption and communication costs. Our experimental results show that the proposed method significantly improves system performance, achieving an accuracy of $92\%$, reducing communication cost by $30\%$, and enhancing client participation compared to traditional FL methods. These results make the proposed approach highly suitable for large-scale edge-cloud computing systems.


\end{abstract}

\section{Introduction}
\label{sec1}

\subsection{Context and Motivation}

\begin{figure}[ht]
    \centering
    \includegraphics[width=.8\textwidth, height=.65\textwidth]{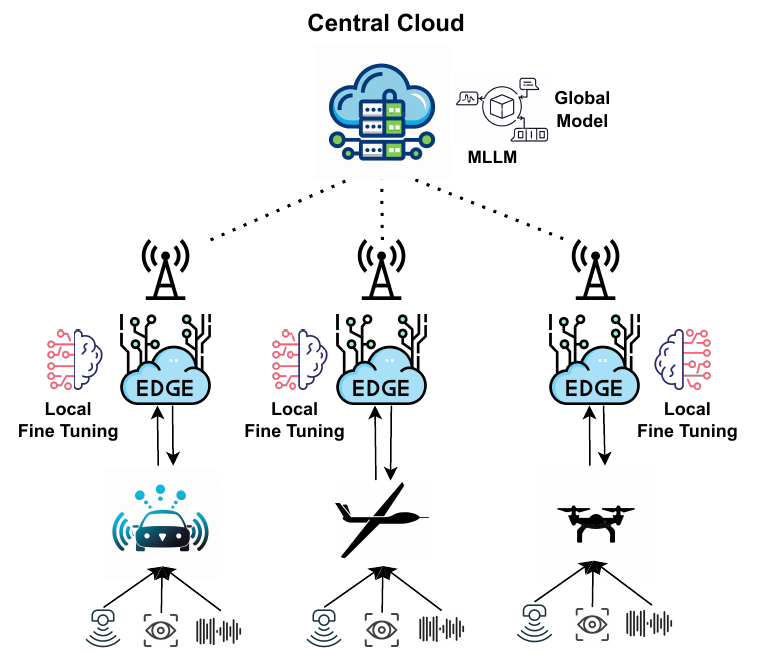}
    \caption{Edge-Cloud-based \textit{MLLM} Architecture with Federated Learning for UAV Networks}
    \label{fig: FL_MLLM_architecture}
\end{figure}

Internet of Things (IoT) devices have become increasingly prevalent, generating vast amounts of data that necessitate real-time processing and intelligent decision-making. To effectively analyze this diverse and extensive data, deep learning approaches have emerged, particularly Multimodal Large Language Models (\textit{MLLMs}), which excel in processing and understanding various data types, including text, images, audio, and sensor readings. These models have contributed significantly to advancements in several emerging domains \cite{huang2024large}, such as autonomous systems \cite{aldeen2024initial}, smart healthcare \cite{wang2023accelerating}, and industrial IoT \cite{fan2024new, usecase_04}. 
In this context, the integration of edge-cloud computing systems with Federated Learning (FL) has garnered substantial attention for its ability to facilitate decentralized training of the deep learning model while ensuring data privacy and security \cite{Akhtarshenas24, bao2022federated, wu2024topology, trindade2022resource,sagar2024hierarchical}. This approach allows for model training directly on IoT devices, thereby preserving sensitive data and enhancing privacy. Fig. \ref{fig: FL_MLLM_architecture} illustrates a general architecture for such integration. In this setup, a central cloud server maintains a global model (\textit{MLLM}), while edge devices perform local fine-tuning for specific tasks (e.g., self-driving cars and drones). Only model updates, rather than raw data, are shared with the central cloud, enhancing the global model while preserving data privacy.

Despite these potential benefits, the deployment of \textit{MLLMs} in edge-cloud systems presents several critical challenges. First, resource constraints on edge devices \cite{sajnani2024secure}, such as limited computational power and battery life, restrict their ability to participate continuously in FL training rounds. Second, the communication overhead associated with transmitting model updates between distributed edge devices and cloud infrastructure can overwhelm networks, particularly in large-scale deployments. Third, the non-IID (non-independent and identically distributed) nature of data across edge devices adds complexity to the model training process, as local data often varies significantly in structure and relevance.

To address the limitations of edge devices, which often lack the necessary resources and capacities for large models like \textit{MLLMs}, we introduce a framework that deploys pretrained \textit{MLLMs} on these devices for fine-tuning, while primary training is conducted in the cloud. In this framework, Particle Swarm Optimization (\textit{PSO}) is employed to efficiently select suitable devices for \textit{MLLM} deployment to ensure that only the most appropriate devices, which have sufficient resources and relevant data, are involved in each FL training round. This selection process helps to tackle the challenges posed by non-IID data of edge devices by providing a more representative sample of the overall data distribution. As a result, it facilitates effective updates to the \textit{MLLM} and enhances its generalization capabilities, even despite the heterogeneity of local data. 
Additionally, Ant Colony Optimization (\textit{ACO}) is employed to optimize communication pathways between the selected edge devices and cloud nodes, facilitating the efficient sharing of model updates \cite{abualigah2023swarm, sun2020survey}. This optimization not only minimizes bandwidth usage but also reduces latency and conserves energy, thereby enhancing the overall performance of the framework. 

The choice of \textit{PSO} and \textit{ACO} is motivated by their proven efficacy in solving complex optimization problems in distributed and dynamic environments. \textit{PSO} is well-suited for optimizing continuous, high-dimensional search spaces, as it efficiently converges towards near-optimal solutions with fewer iterations, making it ideal for edge device selection where multiple criteria must be considered \cite{cao2024computational}. On the other hand, \textit{ACO} excels in discrete optimization problems and has been shown to effectively find optimal or near-optimal paths in dynamic networks by mimicking the natural foraging behavior of ants. This makes \textit{ACO} particularly suitable for optimizing communication pathways in the dynamic and resource-constrained edge-cloud environment, where path quality may change frequently due to varying network conditions \cite{jiang2024evolutionary}.
By combining \textit{PSO} and \textit{ACO}, our hybrid approach aims to balance the computational load across edge devices while ensuring efficient communication between the edge and the cloud. This dual optimization process facilitates the effective deployment of \textit{MLLMs} in edge-cloud computing environments, improving system scalability and performance. In this paper, we evaluate the effectiveness of the proposed hybrid model through a series of experiments conducted in a simulated edge-cloud system. Our results demonstrate significant improvements in energy efficiency, communication overhead, and model accuracy compared to traditional FL approaches.

\subsection{Contributions}

In this paper, we present a novel hybrid swarm intelligence approach to optimize the deployment of \textit{MLLMs} in FL within smart edge-cloud computing systems. Our key contributions are summarized as follows:

\begin{enumerate}
    \item \textbf{Hybrid \textit{PSO}-\textit{ACO} Optimization Framework for Efficient Deployment of \textit{MLLMs}:} We propose a hybrid optimization framework that integrates \textit{PSO} and \textit{ACO}. In this framework, \textit{PSO} is utilized for the efficient selection of edge devices for deploying pretrained \textit{MLLMs}, based on criteria such as resource availability, data relevance, and network stability. \textit{ACO} is employed to optimize communication pathways and facilitate the sharing of \textit{MLLM} updates between the selected edge devices and cloud servers, minimizing communication overhead and latency.

   \item \textbf{Addressing Non-IID Data}: The proposed \textit{PSO}-based method tackles the challenges associated with non-IID data by intelligently selecting a subset of edge devices to ensure diverse contributions to the global model. This approach enhances the generalization and accuracy of the \textit{MLLM} despite the heterogeneity present in the local data distributions.

    \item \textbf{Energy Efficiency and Model Performance Trade-off Optimization:} Our approach balances energy consumption with model accuracy, achieving significant reductions in total energy usage while maintaining or improving the performance of the \textit{MLLM}. This is particularly important for prolonging the operational life of edge devices in resource-constrained environments. 

    \item \textbf{Scalability and Adaptability in Dynamic Environments:} The framework is designed to scale effectively and adapt to real-time variations in edge-cloud environments, managing large numbers of devices and varying network conditions while ensuring optimal system performance.

    \item \textbf{Extensive Experimental Validation:} We conducted comprehensive simulations to evaluate the performance of our hybrid \textit{PSO}-\textit{ACO} framework. The results demonstrate notable improvements in energy efficiency, communication overhead reduction, and model accuracy compared to relevant benchmarks, including deep and reinforcement learning-based approaches.

\end{enumerate}

These contributions collectively provide a robust solution to the challenges of deploying \textit{MLLMs} in FL systems within smart edge-cloud computing environments. Our hybrid \textit{PSO}-\textit{ACO} framework enhances the efficiency, scalability, and performance of FL, making it more viable for real-world applications involving large-scale, resource-constrained edge networks.

The paper is structured as follows: Section \ref{Problem_Formulation} presents the problem formulation, describing the objectives and constraints of the federated learning scenario with the integration of swarm intelligence over the edge-cloud computing environment. Section \ref{Background} provides a brief background on \textit{MLLMs}, including their training process and fine-tuning methods, which are crucial for understanding the implementation of our proposed solution. Section \ref{section2} gives an overview of the related work in FL, swarm intelligence, and edge-cloud computing, establishing the foundation for the proposed approach. Section \ref{section3} details the proposed hybrid \textit{PSO}-\textit{ACO} framework, explaining the integration of particle swarm optimization and ant colony optimization to enhance model performance. The experimental setup and simulation results are presented in Section \ref{results}, where we compare the performance of the proposed framework against relevant benchmarks, including traditional and learning-based models. Finally, Section \ref{section6} concludes the paper by summarizing the findings and discussing potential future research directions.

\section{Problem Formulation} \label{Problem_Formulation}

Deploying \textit{MLLMs} in \textit{FL} systems within \textit{smart edge-cloud computing environments} poses several critical challenges. These challenges arise due to the decentralized nature of FL, the limited resources of edge devices, and the inherent complexities in managing communication and model training across a large and diverse network of devices. In this section, we outline the key challenges that must be addressed to enable the efficient deployment of \textit{MLLMs} in FL systems.

\subsection{Resource Constraints on Edge Devices}

One of the primary challenges in deploying \textit{MLLMs} in FL is the limited computational and energy resources available on edge devices, such as IoT sensors, cameras, and mobile devices. These devices often have restricted battery life and processing power, making it impractical for all devices to participate in every round of FL training. Given the high computational demands of training \textit{MLLMs}, selecting a subset of edge devices that can contribute effectively to the global model without exhausting their resources is essential.

The total energy consumption \( E_i \) for each device \( i \) includes the energy required for both model training \( E_{\text{train}} \), communication \( E_{\text{comm}} \) and time \( E_{\text{time}} \):

\begin{equation}
E_i = E_{\text{train}} + E_{\text{comm}} + E_{\text{time}}
\label{eq1}
\end{equation}

\noindent where \( E_{\text{train}} \) is the energy consumed during training, \( E_{\text{comm}} \) and \( E_{\text{train}} \) represents the energy and time required to transmit model updates to a central server or cloud coordinator, respectively. This central server aggregates local model updates from multiple edge devices. This aggregation step is crucial in FL for updating the global model, which is then shared back with the participating devices.

In FL, training takes place over multiple rounds, where each round consists of three main steps: \textbf{(1)} Global Model Distribution: The central server sends the current global model to a selected set of edge devices. \textbf{(2)} Local Training: Each selected device trains the model on its local data for a specified number of iterations, using its own computational resources. \textbf{(3)} Model Update Communication: The locally updated models are then transmitted back to the central server for aggregation. The process is repeated for several rounds, with the goal of gradually improving the global model's accuracy. The challenge is to minimize the total energy consumption across the selected devices during these rounds, while still ensuring meaningful contributions to the global model updates:

\begin{equation}
\min_{S \subseteq D} \sum_{i \in S} E_i
\label{eq2}
\end{equation}

\noindent where \( S \) represents the subset of devices selected from the entire device set \( D \).

\subsection{Communication Overhead and Latency}

Another significant challenge is the communication overhead associated with transmitting large model updates between edge devices and the cloud or edge servers. In FL, frequent communication is required to synchronize local model updates with the global model, which can result in substantial bandwidth usage, network congestion, and increased latency. This is particularly problematic when deploying \textit{MLLMs}, which typically have large model sizes.

The communication cost \( C_{\text{comm}} \) for each device \( i \) can be modeled as follows:

\begin{equation}
C_{\text{comm}} = \sum_{i \in S} \frac{M_i}{B_i} \cdot d_i
\end{equation}

\noindent where \( M_i \) is the size of the model update, \( B_i \) is the available bandwidth, and \( d_i \) represents the distance between device \( i \) and the nearest server. Reducing communication overhead and minimizing latency is crucial to maintaining the efficiency of the FL process, particularly in large-scale deployments.

\subsection{Non-IID Data Distribution Across Devices}

In FL systems, data is distributed across edge devices in a non-IID (non-independent and identically distributed) manner, meaning that each device collects data that reflects its specific environment or context. This leads to significant variability in the type and distribution of data across the network, which can negatively affect the performance and generalization of the global model, especially when training \textit{MLLMs} that rely on diverse and balanced data.

The local loss function \( \mathcal{L}_i(w) \) for each device \( i \) is computed based on its local data, where \( w \) represents the model parameters:

\begin{equation}
\mathcal{L}_i(w) = \frac{1}{n_i} \sum_{j=1}^{n_i} \ell(f(x_j; w), y_j)
\end{equation}

\noindent where \( n_i \) is the number of local data points, \( \ell(f(x_j; w), y_j) \) is the loss function between the model prediction \( f(x_j; w) \) and the true label \( y_j \). The global objective in FL is to minimize the weighted sum of the local losses across all selected devices:

\begin{equation}
\mathcal{L}(w) = \sum_{i \in S} \frac{n_i}{n} \mathcal{L}_i(w)
\end{equation}

\noindent where \( n \) is the total number of data points across all devices. The challenge here is to ensure that the global model generalizes well despite the heterogeneity in the local data distributions.

\subsection{Energy and Time Efficiency vs. Model Performance Trade-off}

A critical challenge in FL systems is balancing the energy and time consumption of edge devices with the overall performance of the \textit{MLLM}. Edge devices with limited energy and time resources may still hold valuable data that could significantly improve the global model, however, their continuous participation could lead to battery depletion, increased latency, and potential system failures. Conversely, prioritizing solely on energy and time efficiency may result in the exclusion of devices with high-quality data essential for effective model training. Balancing these factors is crucial to ensuring both sustainable resource usage and optimal model performance across the network.

The total energy consumed by a device can be formulated as shown in Eq.\ref{eq1}.
The optimization problem involves minimizing a weighted combination of energy consumption and model performance as follows:

% \begin{equation}
% \min_{S \subseteq D} \alpha \sum_{i \in S} E_i + \beta \mathcal{L}(w),
% \end{equation}

\begin{equation}
\min_{S \subseteq D} \alpha \sum_{i \in S} E_i + \gamma \sum_{i \in S} T_i + \beta \mathcal{L}(w) 
\end{equation}

\noindent where \( \alpha \), \( \beta \), and \( \gamma \) are weighting factors that balance the trade-off between energy efficiency and model accuracy.

\subsection{Scalability and Adaptability in Dynamic Edge-Cloud Environments}

As edge-cloud systems scale to include thousands or even millions of devices, the need for scalability and dynamic adaptability becomes increasingly important. Edge devices are often mobile and subject to changing environmental conditions, leading to variability in resource availability and connectivity. This dynamic nature requires FL systems to continuously adapt to changes in the network, ensuring efficient use of resources and maintaining high model performance.

Scalability challenges arise in ensuring that the system can manage the increasing number of devices without overwhelming network resources or introducing significant delays. Dynamic adaptability involves efficiently handling devices joining or leaving the network, adjusting communication paths, and reallocating computational tasks in real time.

\section{MultiModal Large Language Model (MLLM) Background}\label{Background}

\begin{figure}[htbp]
    \centering
    \includegraphics[width=\textwidth, height=.28\textwidth]{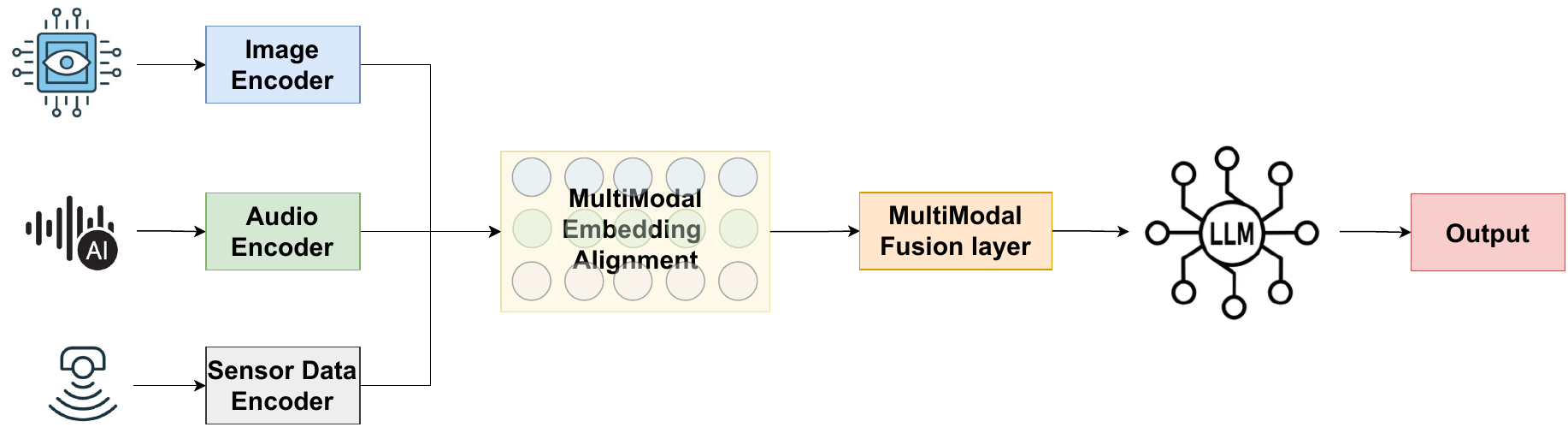}
    \caption{Architectural Overview of \textit{MLLM}. }
    \label{fig:MLLM_architecture}
\end{figure}

\textit{MLLM} are advanced machine learning models capable of processing and integrating data from multiple modalities, such as text, images, audio, and structured sensor data \cite{MLLM_02}. The structure and architecture of \textit{MLLM} center on integrating a Large Language Model (\textit{LLM}) with additional components to handle other data modalities, like images, audio, or video. \textit{MLLMs} primarily have a language-centric architecture where other modalities are treated as extensions, allowing the model to understand and generate responses based on complex, multimodal input \cite{MLLM_01}. Fig. \ref{fig:MLLM_architecture} depicts the architectural overview of \textit{MLLMs} illustrating core components and workflow.

\subsection{Backbone of MLLM}

\textbf{LLM Foundation:} The core of an \textit{MLLM} is an \textit{LLM} such as GPT, PaLM, or BERT, trained on extensive text datasets \cite{MLLM_00}. The language model is typically a transformer-based model with self-attention mechanisms that enable it to understand and generate text \cite{Transformer}.\\

\textbf{Pretrained Text Understanding:}  \textit{MLLMs} leverage the language model’s strong natural language understanding and generation capabilities. This foundation ensures that the model can perform text-based tasks and serves as the central processing unit for interpreting other modalities \cite{MLLM_03}.

\subsection{Modality-Specific Encoders}

To handle non-textual data like images or audio, modality-specific encoders are used. The encoders transform each modality into embeddings that can be processed by the language model \cite{MLLM_modality}.

\begin{itemize}
    \item \textbf{Image Encoder}: Given an image \( I \), it is divided into patches, and each patch is flattened into a vector \cite{MLLM_image}. These patch vectors are linearly embedded into a vector space, similar to text token embeddings:
    \begin{equation}
    f_i = \text{Embed}(\text{patch}_i) \quad \text{for each patch}_i
    \end{equation}
    This results in image embeddings \( F = \{f_1, f_2, \ldots, f_m\} \), where each \( f_i \in \mathbb{R}^d \).

    \item \textbf{Audio Encoder}: Audio data \( A \) is processed into a sequence of feature vectors $a_j$ using an encoder like wav2vec \cite{MLLM_audio}.
    \begin{equation}
    a_j = \text{wav2vec}(A_j) \quad \text{for each audio segment}~j
    \end{equation}
    This results in audio embeddings \( A = \{a_1, a_2, \ldots, a_k\} \), with each \( a_j \in \mathbb{R}^d \).

    \item \textbf{Sensor Data Encoder}: Sensor data \( S \) is processed into a sequence of feature vectors $S_j$ using an encoder module of Transformer model \cite{Transformer}.
    \begin{equation}
    S_z = \text{Transformer-Encoder}(S_z) \quad \text{for each sequential data segment}~z
    \end{equation}
    This results in audio embeddings \( S = \{s_1, s_2, \ldots, s_k\} \), with each \( S_z \in \mathbb{R}^d \).
\end{itemize}

\subsection{Multimodal Embedding Layer}

The outputs of the modality-specific encoders (e.g., image and audio embeddings) are mapped to a unified embedding space compatible with the language model. These embeddings are typically augmented with special tokens to identify their modality \cite{MLLM_embadding}.

Let \( E_{\text{text}} \) represent text embeddings, \( E_{\text{image}} \) represent image embeddings, \( E_{\text{audio}} \) represent audio embeddings. Each modality is projected into the shared space by learned projections:
\begin{equation}
E'_{\text{modality}} = W_{\text{modality}} \times E_{\text{modality}}
\end{equation}
where \( W_{\text{modality}} \) is a learnable weight matrix that aligns each modality to the language model’s embedding space.

\subsection{Fusion Mechanism}

\begin{itemize}
    \item \textbf{Attention-Based Fusion}: 
    \textit{MLLMs} typically leverage the transformer’s self-attention mechanism to fuse multimodal information. During processing, each token (whether from text, image, or audio) attends to every other token, allowing the model to integrate information across modalities. This attention mechanism enables the model to focus on relevant aspects of each modality for the task at hand. The multimodal tokens (from text, image, audio) are combined in a single self-attention mechanism, allowing all tokens to interact \cite{MLLM_attention}.
    \begin{equation}
    h_{\text{multi}}^{(l+1)} = \text{Attention}(Q_{\text{multi}}, K_{\text{multi}}, V_{\text{multi}})
    \end{equation}
    This fused representation \( h_{\text{multi}} \) integrates information from all modalities.
    
    \item \textbf{Cross-Attention Layers}: In some \textit{MLLMs}, additional cross-attention layers are introduced to explicitly integrate visual, audio, or other embeddings into the language model. Cross-modal attention directs the model to selectively focus on the most relevant information from non-text modalities in context with the text. For cross-attention, each modality (e.g., text, image) has separate attention layers to capture relationships \cite{MLLM_cross}.
    \begin{equation}
    \text{Attention}(Q_{\text{lang}}, K_{\text{modality}}, V_{\text{modality}}) = \text{softmax}\left(\frac{Q_{\text{lang}} K_{\text{modality}}^\top}{\sqrt{d_k}}\right) V_{\text{modality}}
    \end{equation}
    where \( Q_{\text{lang}} \) represents the language model’s query vectors, and \( K_{\text{modality}} \) and \( V_{\text{modality}} \) are the key and value vectors from the other modality.
\end{itemize}

\subsection{Modality-to-Text Generation}

Once all modalities are aligned in the embedding space, the language model backbone generates text outputs based on multimodal context \cite{MLLM_align}.

\begin{itemize}
    \item \textbf{Language Modeling Objective}: The language model produces text by predicting the next token \( y_t \) given previous tokens and the multimodal context \cite{MLLM_align_lang}.
    \begin{equation}
    P(y_t | y_{<t}, h_{\text{multi}})
    \end{equation}
    where \( h_{\text{multi}} \) represents the fused multimodal embedding.

    \item \textbf{Text Output Generation}: Using the language model’s autoregressive nature, it generates a sequence of tokens \( \{y_1, y_2, \ldots, y_T\} \), forming a coherent output based on multimodal inputs \cite{MLLM_align_generation}.
\end{itemize}

\subsection{Training Process}

Training of \textit{MLLMs} typically involves two stages.

\begin{itemize}
    \item \textbf{Text Pretraining}: The language model is pre-trained on a large corpus of text to establish language understanding, optimizing \cite{MLLM_pretrain}.
    \begin{equation}
    \min \mathcal{L}_{\text{text}} = -\sum_{t} \log P(y_t | y_{<t})
    \end{equation}

    \item \textbf{Multimodal Fine-Tuning}: After text pretraining, the model is fine-tuned on multimodal datasets with paired text and other modalities, optimizing for cross-modal tasks \cite{MLLM_finetune}.
    \begin{equation}
    \min \mathcal{L}_{\text{multi}} = -\sum_{t} \log P(y_t | y_{<t}, h_{\text{multi}})
    \end{equation}
    The objective is to maximize the likelihood of generating correct responses based on multimodal context.
\end{itemize}

\section{Related Work}
\label{section2}
In this section, we examine the relevant research on the deployment of \textit{MLLMs}, focusing on FL and edge systems. We will also discuss related studies on the integration of \textit{MLLMs} with swarm intelligence.

\subsection{Deployment of MLLMs and LLMs}
\textit{MLLMs} and \textit{LLMs} are extensively used in the AI community for their ability to process enormous amounts of data. Their foundation on attention mechanisms enables them to efficiently handle long data sequences while preserving dependencies within the data. Notable \textit{LLMs} in use currently include BERT \cite{tang2023fusionai, mudvari2024splitllm, jiang2024low, liu2023differentially}, GPT and its variants \cite{tang2023fusionai, mudvari2024splitllm, kou2024pfedlvm, liu2023differentially}, CLIP \cite{atapour2024leveraging, zhang2024mllm}, and various LLAMA models \cite{dong2024fine, zhao2024llm, liu2023differentially, ye2024openfedllm, wu2024fedbiot, kuang2024federatedscope}, among others. However, challenges related to scalability and privacy remain major obstacles when deploying \textit{LLMs}, particularly \textit{MLLMs} \cite{krishnamoorthy2024integrating, kurkute2023scalable}. Federated learning is one of the emerging solutions to address this issue, as it allows \textit{MLLMs} to be deployed on edge devices that collaboratively train a shared model without exchanging their local data \cite{kurkute2023scalable,yao2024federated,ye2024openfedllm}. However, \textit{MLLMs} are known for their high memory and computational demands, making deployment particularly challenging on edge devices with limited consumer-level GPUs. 

Various approaches have been proposed to address the issue of computing resources, aiming to enhance the flexibility and reusability of \textit{MLLMs}/\textit{LLMs} while ensuring secure and private data handling. For instance, solutions like task scheduling or organizing machine learning tasks have been explored to facilitate this process \cite{tang2023fusionai}, along with solutions that utilize model compression techniques and Parameter-Efficient Fine-Tuning methods \cite{dong2024fine,jiang2024low,liu2023differentially,kim2023efficient,wu2024fedbiot,kuang2024federatedscope}.
%like LoRA (Low-Rank Adaptation) , smaller-sized adapters %. 
At the same time, other approaches have been proposed to avoid deploying resource-intensive models on edge devices and instead use alternatives that effectively leverage these powerful models in constrained environments. For example, in the context of federated learning systems, the authors in \cite{atapour2024leveraging} propose using Knowledge Distillation and a Prompt Generator to generates knowledge from the combined data of multiple IoT devices, allowing the model to be updated without the need for full deployment on edge devices, thereby keeping sensitive data private. Similarly, in \cite{zhang2024mllm}, the authors propose pretraining \textit{LLMs} on the server before fine-tuning them on edge devices. This approach aims to minimize both computational resource usage and memory consumption related to deploying large models on edge devices. In a related work \cite{mudvari2024splitllm}, the authors also suggest dividing the computational tasks of \textit{LLMs} between the server and edge devices to address the limited capabilities of these devices, while another work \cite{zhao2024llm} suggests distributing sensitive layers of the \textit{LLM} on client devices while offloading non-sensitive layers to the server.

The deployment of large models is not the sole issue, the communication between edge devices and the cloud server also presents challenges. To address the communication overhead, the authors in \cite{kou2024pfedlvm} suggest deploying the large visual model on the cloud server and allowing vehicles to share only the learned features instead of the entire model parameters. This strategy enables each vehicle to keep its training data local while simultaneously reducing communication demands.
Given the ongoing efforts to manage the deployment of \textit{MLLMs}/\textit{LLMs} in decentralized systems like federated learning, including approaches that involve server deployment or sharing layers and data between edge devices and servers, there is still a lack of effective strategies for edge device selection and communication optimization. Therefore, to address this gap, we propose deploying pre trained \textit{MLLMs} on edge devices, where the models are initially trained in the cloud and subsequently fine-tuned on the edge. 

While various approaches in the literature have explored supervised learning, unsupervised learning, and reinforcement learning to optimize federated learning processes \cite{mattoo2024device, eid2024federated, rjoub2021improving, muhammad2020deep, aradi2020survey, rjoub2024trust}, these methods face challenges in handling communication overhead, non-IID data, and the dynamic nature of the environment. Swarm intelligence, with its decentralized structure, ability to select devices based on available resources and data relevance, and capability to optimize communication between edge devices and the cloud, presents a promising solution to these issues. However, there is limited research on the application of swarm intelligence techniques for device selection in federated learning \cite{xing2023efficient, supriya2023particle}. In this paper, we propose leveraging swarm intelligence techniques, including \textit{PSO} and \textit{ACO}, to optimize the deployment strategy of MLLMs, focusing on improving the efficiency of model updates between the edge and the cloud, while considering the resource availability and data relevance of edge devices.

\subsection{MLLMs/LLMs and Swarm Intelligence}
Swarm intelligence is an emerging approach in the field of AI that leverages individual collaboration to achieve a common and overarching goal. In the context of \textit{LLMs}, swarm intelligence is increasingly being used to optimize the performance of these models through multi-\textit{LLM} collaboration. For instance, the authors of this work \cite{feng2024model} have developed a swarm model inspired by \textit{PSO} that enables multiple \textit{LLMs} to work together. They explore the weight space based on successful training checkpoints to enhance performance and make the models easily adaptable to various tasks.
In a different context, the authors in \cite{han2024swarm} have developed an approach based on swarm intelligence and visual question answering mechanisms to enable multiple Large Vision-Language Models (LVLMs) to collaborate on the geo-localization task. This solution allows for linking images with specific geographic locations without requiring a large database of geo-tagged images. At the same time, it leverages the network retrieval capabilities of multiple models, allowing them to collaborate and share knowledge effectively.
Given the extensive variety of swarm intelligence algorithms, practitioners often find it difficult to select the most suitable method for specific tasks, such as designing new metaheuristic algorithms \cite{van2024llamea,pluhacek2023leveraging}. To address this challenge, the authors in \cite{pluhacek2023leveraging} propose a study that uses a \textit{LLM} like GPT-4 to assist in the design process of novel metaheuristics through the hybridization of various swarm intelligence algorithms. However, the authors also highlight several challenges associated with the use of \textit{LLMs}, including their inability to consistently produce accurate and reliable outputs, necessitating human intervention. Additionally, they address the ethical and social concerns that arise from the application of these models.

From our literature review, we found examples of using swarm intelligence to enhance the collaborative use of \textit{LLMs}, making them more adaptable to various tasks or improving their efficiency for specific applications. Additionally, there are instances where \textit{LLMs} are employed to optimize the performance of swarm intelligence algorithms, such as in the design of metaheuristics. However, we noted a lack of research focused on applying swarm intelligence to improve the deployment of \textit{LLMs}, as well as a notable scarcity of studies addressing \textit{MLLMs}. Therefore, in this paper, we propose utilizing hybrid swarm intelligence, combining \textit{PSO} and \textit{ACO}, to optimize the deployment of \textit{MLLMs} specifically in federated learning systems within smart edge-cloud computing environments. Our approach aims to facilitate better resource allocation, improve communication between models, and optimize the overall performance of \textit{MLLMs} in dynamic environments.

\section{Proposed Framework} 
\label{section3}

In this section, we present the proposed hybrid \textit{PSO-ACO} framework, designed to optimize the deployment of \textit{MLLMs} in \textit{FL} environments within smart edge-cloud systems. The framework addresses three critical optimization tasks: \textbf{(1)} selecting the most suitable subset of edge devices for participation in FL rounds, \textbf{(2)} minimizing the communication overhead by optimizing the paths through which model updates are transmitted to cloud servers, and \textbf{(3)} handling the Non-IID nature of data across devices. The model leverages \textit{PSO} for edge device selection and \textit{ACO} for communication path optimization.

\subsection{Overview of the Hybrid \textit{PSO-ACO} Framework}

Swarm intelligence is an evolving approach that leverages individual collaboration of agents to achieve a common goal. In the context of \textit{MLLMs}, swarm intelligence has begun to play a significant role in optimizing \textit{MLLM} performance through collaborative strategies. In this work, a hybrid framework based on two swarm intelligence techniques, \textit{PSO} and \textit{ACO}, is proposed to optimize the deployment of \textit{MLLMs} in FL edge-cloud environments. This hybrid approach is organized into three main stages: 

\begin{itemize} 
    \item \textbf{Edge Device Selection via \textit{PSO}}: In each FL round, \textit{PSO} selects a subset of edge devices, each equipped with a pretrained \textit{MLLM}, based on their available resources, data relevance, and energy constraints. This ensures that only devices capable of contributing meaningfully to the global model while minimizing energy consumption are chosen and mitigate the impact of Non-IID data distributions across devices.
      
    \item \textbf{Communication Path Optimization via \textit{ACO}}: Following device selection, each chosen device fine-tunes its corresponding \textit{MLLM}. Then, \textit{ACO} is used to optimize the communication of these fine-tuned model updates between the selected edge devices and the cloud, aiming to minimize communication latency and bandwidth usage to ensure efficient transmission.
    
    \item \textbf{Global Model Update}: The aggregated updates of the fine-tuned \textit{MLLMs} from all selected devices are sent to the cloud through optimized communication paths to update the global \textit{MLLM}.
\end{itemize}

\begin{figure}[htbp]
    \centering
    \includegraphics[width=\textwidth]{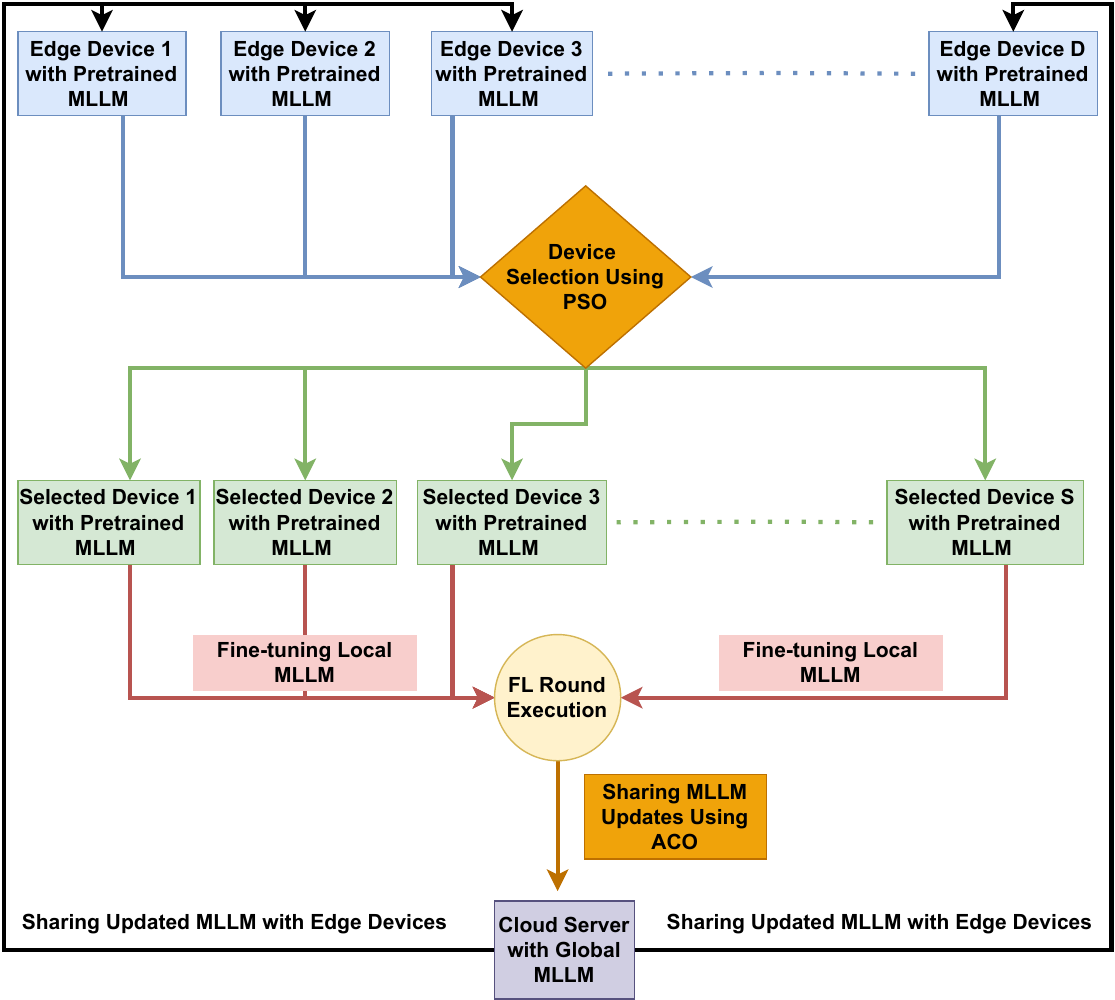}
    \caption{Overview of the Proposed Framework for Optimizing \textit{MLLM} Deployment Using Swarm Intelligence Techniques \textit{PSO} and \textit{ACO}}
    \label{fig:uvs_architecture}
\end{figure}

Each stage is essential for ensuring the efficient deployment of \textit{MLLMs} in resource-constrained, distributed edge-cloud environments.
The proposed solution, outlining the three stages, is shown in Fig.  \ref{fig:uvs_architecture}.

\subsection{\textit{PSO} for Edge Device Selection} 

\textit{PSO} is a population-based stochastic optimization technique inspired by the social behavior of birds flocking or fish schooling. In \textit{PSO}, a group of particles (potential solutions) explores the search space by adjusting their positions based on their own experience and that of neighboring particles. Each particle has a position and a velocity, which are iteratively updated to find the optimal solution. The updates are influenced by the particle's best-known position and the global best position found by the entire swarm, leading the swarm to converge towards optimal or near-optimal solutions.

The first step in the proposed framework is to select the optimal subset of edge devices using \textit{PSO}. The goal is to minimize energy consumption and ensure that selected devices contribute valuable local updates to the global model.
Each particle in the \textit{PSO} algorithm represents a potential solution, i.e., a subset of devices \( S \subseteq D \). The fitness function is designed to minimize the energy consumption \( E_i \) of the selected devices while maximizing the relevance of their local data for \textit{MLLM} fine-tuning. The fitness function is defined as follows:
 \begin{equation}
   \text{Fitness}(S) = \alpha \sum_{i \in S} E_i + \beta \sum_{i \in S} (1 - R_i) + \gamma \sum_{i \in S} D_i
   \label{eq:fitness}
   \end{equation}

\noindent where \( E_i \) represents the energy consumption of device \( i \), \( R_i \) denotes the relevance of the local data to the global model, and \( D_i \) measures the data diversity contributed by device \( i \). The weighting factors \( \alpha \), \( \beta \), and \( \gamma \) balance the trade-offs between energy efficiency, data relevance, and diversity, respectively. The velocity and position of each particle are updated based on the standard \textit{PSO} update equations:

\begin{equation}
v_i(t+1) = \omega v_i(t) + c_1 r_1 (p_i - x_i(t)) + c_2 r_2 (g_i - x_i(t))
\label{eq8}
\end{equation}

\begin{equation}
x_i(t+1) = x_i(t) + v_i(t+1)
\label{eq9}
\end{equation}

\noindent where \( v_i(t) \) is the velocity of particle \( i \) at time $t$, \( x_i(t) \) is its current position of particle \( i \), \( p_i \) is its personal best position for particle \( i \), \( g_i \) is the global best position of particle \( i \), $\omega$ is the inertia weight controlling the influence of the previous velocity, \( c_1, c_2 \) are acceleration coefficients, and \( r_1, r_2 \)  are random values between $0$ and $1$. The algorithm converges to an optimal subset of devices that can participate in the FL round.

Algorithm \ref{algpso} applies \textit{PSO} to select the most appropriate subset of edge devices for participation in a distributed task. The selection is based on both energy consumption and the relevance of the data on each device. By optimizing for these factors, the algorithm ensures that only the most suitable edge devices are selected to participate in a given task, balancing energy efficiency with task relevance.

\begin{algorithm}[H]
\caption{\textit{PSO} for Edge Device Selection}
\label{algpso}
\begin{algorithmic}[1]
\State \textbf{Input:} Set of devices \( D \), energy thresholds \( E_{\text{threshold}} \), relevance metrics \( R_i \)
\State \textbf{Output:} Selected subset of devices \( S \)
\State Initialize particles representing different subsets of devices
\State Initialize velocities and positions of particles
\For{each iteration}
    \For{each particle}
        \State Calculate fitness of each particle based on energy and data relevance
        \State Update particle velocity and position using Equations (\ref{eq8}) and (\ref{eq9})
    \EndFor
    \State Update the global best particle
\EndFor
\State Return the optimal subset of devices \( S \)
\end{algorithmic}
\end{algorithm}

\subsection{\textit{ACO} for Communication Path Optimization}

\textit{ACO} is a nature-inspired metaheuristic algorithm that mimics the foraging behavior of ants to solve complex optimization problems, particularly in finding optimal paths. In nature, ants deposit a chemical substance called pheromone along their paths while searching for food. Other ants tend to follow paths with higher pheromone concentrations, reinforcing those paths over time, which eventually leads to the discovery of the shortest or most efficient route to the food source.

Once the optimal subset of edge devices is selected, \textit{ACO} is applied to optimize the communication paths between the selected edge devices and the cloud server. The primary objective is to minimize both communication latency and bandwidth usage, ensuring efficient transmission of model updates in a distributed edge-cloud environment.

In the \textit{ACO} algorithm, each ant represents a potential communication path between an edge device and the cloud server. The decision-making process for each ant is influenced by two key factors:
(1). Pheromone levels \( \tau_{ij} \): A higher pheromone level indicates a better communication path based on previous ants’ experiences; 
(2). Heuristic information \( \eta_{ij} \): This represents problem-specific knowledge such as path distance or bandwidth, guiding the ants towards more efficient paths.

The probability \( P_{ij} \) of an ant selecting edge \( (i,j) \) is computed using a combination of the pheromone levels and heuristic information:

\begin{equation}
P_{ij} = \frac{\tau_{ij}^\alpha \cdot \eta_{ij}^\beta}{\sum_{k \in \text{neighbors}} \tau_{ik}^\alpha \cdot \eta_{ik}^\beta}
\end{equation}

\noindent where \( \tau_{ij} \) is the pheromone level on edge \( (i,j) \), \( \eta_{ij} \) is the heuristic information, such as bandwidth or distance on edge \( (i,j) \), \( \alpha \) and \( \beta \) are parameters that control the influence of the pheromone level and heuristic information, respectively.

The heuristic information \( \eta_{ij} \) can be defined based on the characteristics of the communication path: if distance or latency is the critical factor, \( \eta_{ij} \) is inversely proportional to the distance \( d_{ij} \) between nodes \( i \) and \( j \):
  \[
  \eta_{ij} = \frac{1}{d_{ij}}
  \]
where shorter distances or lower latencies are preferred.

Alternatively, if bandwidth is the primary factor for path selection, the heuristic information \( \eta_{ij} \) can be expressed in terms of the available bandwidth \( B_{ij} \) on the communication link between nodes \( i \) and \( j \). In this case, \( B_{ij} \) refers to the maximum data transmission capacity of the link, typically measured in megabits per second (Mbps) or gigabits per second (Gbps). To model the influence of bandwidth on path selection, the heuristic information is directly proportional to the available bandwidth:

\[
\eta_{ij} = \frac{1}{B_{ij}}
\]

\noindent where \( B_{ij} \) is the available bandwidth on the communication link between nodes \( i \) and \( j \), paths with smaller values of \( B_{ij} \) (i.e., lower bandwidth) will result in larger \( \eta_{ij} \), making those paths less likely to be chosen by the ants.

This inverse relationship ensures that paths with higher bandwidth are assigned lower heuristic values \( \eta_{ij} \), thereby encouraging the selection of paths that can support greater data transmission rates. The incorporation of bandwidth into the heuristic information guides the ants towards paths that are more efficient for communication, improving overall network throughput and reducing congestion.

The pheromone levels \( \tau_{ij} \) are updated after each iteration to reinforce better communication paths. The update equation for \( \tau_{ij} \) is:

\begin{equation}
\tau_{ij}(t+1) = (1-\rho) \tau_{ij}(t) + \Delta \tau_{ij}
\end{equation}

\noindent where \( \rho \) is the pheromone evaporation rate, which ensures that previously chosen paths are not overly reinforced, allowing the algorithm to explore new paths.
\( \Delta \tau_{ij} \) is the pheromone deposit, which is proportional to the quality of the path based on factors such as latency or bandwidth.

\begin{algorithm}[H]
\caption{\textit{ACO} for Communication Path Optimization}
\label{algaco}
\begin{algorithmic}[1]
\State \textbf{Input:} Selected devices \( S \), pheromone levels \( \tau_{ij} \), heuristic \( \eta_{ij} \)
\State \textbf{Output:} Optimized communication paths
\For{each ant}
    \State Construct a path from device \( i \) to the cloud using probability \( P_{ij} \)
    \State Evaluate the path based on bandwidth usage and latency
    \State Update pheromone levels on path edges based on path quality
\EndFor
\State Return the optimal communication paths
\end{algorithmic}
\end{algorithm}

The \textit{ACO} algorithm (Algorithm \ref{algaco}) iterates through a number of devices, each exploring different communication paths between the selected edge devices and the cloud. Paths that offer higher bandwidth and lower latency are favored, and over successive iterations, the devices collectively converge on the most efficient paths. This leads to an optimized communication network that minimizes both bandwidth usage and latency, ensuring that model updates are transmitted efficiently and effectively.

\subsection{Global Model Aggregation and Update}

After local model fine-tuning is completed on the selected edge devices, the optimized communication paths are used to transmit the local updates to the cloud. The cloud server aggregates the local updates to form the global model. The aggregation process is defined as follows:

\begin{equation}
w_{\text{global}} = \sum_{i \in S} \frac{n_i}{n} w_i
\label{eq12}
\end{equation}

\noindent where \( w_i \) is the local model update from device \( i \), \( n_i \) is the number of data samples on device \( i \), and \( n \) is the total number of data samples across all devices. The global model is then updated, and the process repeats for subsequent FL rounds.

\section{Experimental Setup and Evaluation}
\label{results}

This section presents the experiments carried out to assess the proposed hybrid framework, including a detailed description and analysis of the results obtained. For the purposes of this work, the Unmanned Vehicle System (UVS) was selected as the use case.

\subsection{Use Case Scenario \& operational flow}

UVSs are autonomous robotic vehicles that function without the need for onboard human control. These systems are capable of performing various tasks, such as navigation and surveillance, in both terrestrial and aerial domains, and have seen recently significant progress in terms of technological development \cite{dong2024communication}. 
A typical UVS is equipped with several sensors that collect diverse types of data, such as images, LIDAR, audio, IMU, and GPS data. These sensors may consist of:
    \begin{itemize}
        \item \textit{Cameras}: Capture real-time images and videos used for obstacle detection and lane following.
        \item \textit{LIDAR Sensors}: Generate 3D maps for object distance measurement and spatial awareness.
        \item \textit{Microphones}: Collect environmental audio cues and voice commands.
        \item \textit{IMU}: Tracks the vehicle’s orientation, speed, and acceleration.
        \item \textit{GPS}: Provides precise geolocation data for navigation and route planning.
    \end{itemize}

This variety of sensors generates large quantities of data, which require sophisticated models for the processing of extensive, real-time multimodal data. To enable optimized resource selection and efficient communication for decision-making in the UVS, a well-designed processing architecture is essential. In this context, \textit{MLLM} is particularly well-suited due to its robustness and advanced capabilities.

As discussed in Section \ref{Background}, \textit{MLLMs} integrate different data types through a language-centric framework that seamlessly incorporates other modalities, such as images, audio, and structured sensor data. This integration enables \textit{MLLMs} to generate responses and make decisions based on complex multimodal inputs, like those provided by UAVs. By leveraging these advanced architectures, UVSs can achieve real-time processing and efficient data integration, optimizing decision-making for tasks such as navigation, object recognition, and environmental awareness.

%A comprehensive use case scenario has been depicted for an \textit{UVS} that integrates various cutting-edge technologies, including multimodal data processing using a \textit{MLLM}, \textit{Swarm Intelligence} (PSO and ACO), and \textit{Edge-Cloud Computing} in the IoT environment, \textcolor{red}{figure [ ]}. This architecture requires real-time data processing, optimized resource selection, and efficient communication for decision-making in the UVS \cite{usecase_00}.
% \subsection{ of the UVS}

% The operational flow of the UVS begins with the collection of multimodal data from the various sensors. This data is then processed in parallel by the respective modules (Transformer-based vision and language models, Point Cloud, IMU Neural Networks), and the resulting features are fused in the Feature Fusion Layer of the MLLM. Next, the PSO algorithm selects the edge devices that are most suitable for real-time data processing based on their computational resources and energy levels. Then, ACO optimizes the communication paths between the selected edge devices and the cloud, minimizing latency and bandwidth usage. Finally, the fused data is used for real-time decision-making, such as obstacle avoidance, lane following, and route planning. The UVS continuously communicates with the cloud for global model updates and enhanced decision-making capabilities.

In the \textit{MLLM}-driven operational flow of the UVS, the process, as presented in Section \ref{Background}, starts with the collection of multimodal data from various sensors. This multimodal data is then independently processed by specialized modules within the \textit{MLLM} framework: transformer-based models handle vision and language data, while point cloud and IMU data are processed by dedicated neural networks \cite{usecase_01}. Each module extracts essential features from its modality, which are then fused in the \textit{MLLM}’s Feature Fusion Layer to create a comprehensive, unified representation of the environment \cite{usecase_multimodal}. 

\begin{figure}[ht]
    \centering
    \includegraphics[width=\textwidth, height=.53\textwidth ]{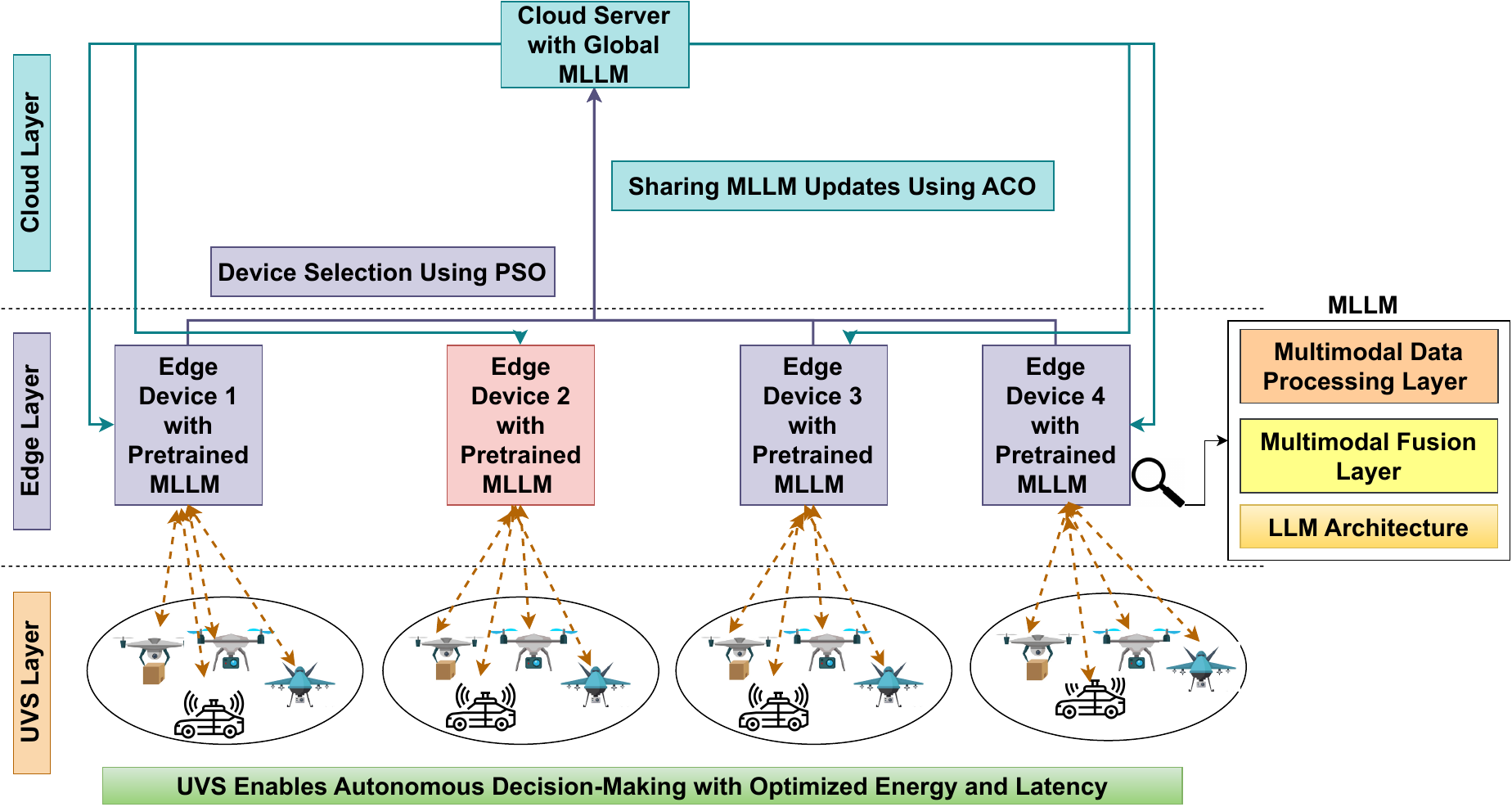}
    \caption{The Architecture of the Use Case Solution}
    \label{fig:uvs_architecture1}
\end{figure}

In this scenario, the UVS functions within a hybrid edge-cloud computing framework where the local data processing occurs on the edge devices, while global model updates and more complex computations are handled in the cloud. 
To optimize the deployment of \textit{MLLM} on edge devices, \textit{PSO} is employed to select the most appropriate devices based on factors such as resource availability (e.g., battery life, computational capacity) and data relevance. This ensures that the most capable devices contribute to the real-time processing tasks of the UVS.
Once the devices are selected, the fine-tuned \textit{MLLM} models are transferred to the cloud server. At this stage, \textit{ACO} is used to optimize the communication paths between the selected edge devices and the cloud. The objective is to minimize communication latency and bandwidth consumption, ensuring the efficient transmission of model updates and real-time data.\cite{usecase_ACO,usecase_01, usecase_02}.
Following this architecture, the fused multimodal representation, processed and distributed across optimal pathways, enables the \textit{MLLM} to support real-time decision-making tasks directly at the edge such as obstacle avoidance, lane following, and route planning. Continuous communication with the cloud allows the \textit{MLLM} to access global model updates, which further enhances the UVS’s ability to make accurate and adaptive decisions in dynamic environments \cite{usecase_03,usecase_edge}.
The architecture of the use case solution is illustrated in Fig. \ref{fig:uvs_architecture1}.

%\begin{figure}[h!]
%    \centering
%    \includegraphics{UVS.drawio.pdf}
%    \caption{The architecture of the use case solution}
%    \label{fig:uvs_architecture1}
%\end{figure}

\subsection{Simulation Environment}

To evaluate the effectiveness of the UVS model, we designed a simulation environment that closely mimics real-world urban scenarios. This environment includes multiple edge devices with diverse computational capacities and battery levels, reflecting the heterogeneous nature of real-world IoT networks. Additionally, the cloud infrastructure was simulated to assess the edge-cloud communication efficiency under various conditions.
The Key parameters of the simulation environment are as follows:
\begin{itemize}
    \item \textbf{Number of Edge Devices:} $50$ to $200$ devices, simulating onboard and nearby devices in a vehicular network.
    \item \textbf{Edge Device Capabilities:} Devices have varying levels of processing power (CPU, GPU) and battery life.
    \item \textbf{Cloud Infrastructure:} A centralized cloud server for model updates and complex computations.
    \item \textbf{Communication Range:} $100$ to $500$ meters between edge devices, depending on the simulation scenario.
    \item \textbf{Bandwidth:} Varying bandwidths between $10$ and $100$ Mbps for different communication channels.
\end{itemize}

\subsubsection{Datasets: Multimodal Data Sources}

The \textit{Leddar PixSet} dataset \footnote{https://leddartech.com/solutions/leddar-pixset-dataset/} is used for LIDAR, camera, radar, IMU, and GPS data. Since the dataset does not provide audio data, we complement it with the UrbanSound8K dataset \footnote{https://urbansounddataset.weebly.com/urbansound8k.html}, which includes high-quality audio recordings. This combination of datasets enables full multimodal testing of the UVS system.

\begin{itemize}
    \item \textbf{Leddar PixSet Dataset:} This dataset contains the following information:
    \begin{itemize}
        \item \textbf{LIDAR:} Captures 3D point cloud data for object detection and spatial mapping.
        \item \textbf{Cameras:} Provides RGB images for obstacle detection, lane tracking, and object classification.
        \item \textbf{Radar:} Adds depth perception and robustness in poor weather conditions.
        \item \textbf{IMU:} Provides inertial data to estimate orientation, acceleration, and vehicle dynamics.
        \item \textbf{GPS:} Supplies precise geolocation data for real-time navigation and path planning.
    \end{itemize}

    \item \textbf{UrbanSound8K Dataset:} This dataset provides labeled sound excerpts for common urban sounds, including vehicle horns, engine noises, emergency sirens, and street noise. It is ideal for testing environmental sound recognition within the UVS.
\end{itemize}

The Leddar PixSet dataset offers over $29,000$ frames and $97$ sequences in urban traffic scenarios, while the UrbanSound8K dataset provides thousands of real-world urban sound recordings, allowing for the evaluation of the UVS system across both visual, spatial, inertial, and auditory modalities.

\subsubsection{Baseline Methods}

%\begin{table}[H]
%  \centering
%  \begin{tabular}{|p{4.0cm}|p{6.0cm}|}
%\hline
%\rowcolor{red!30}
%   LMM/LLM & References \\ \hline

% BERT & \cite{tang2023fusionai,mudvari2024splitllm,jiang2024low,liu2023differentially}  \\
%\hline

% GPT \& Variants & \cite{tang2023fusionai,mudvari2024splitllm,kou2024pfedlvm,liu2023differentially}  \\
%\hline

% CLIP & \cite{atapour2024leveraging,zhang2024mllm}  \\
%\hline

% LLAMA \& Variants & \cite{dong2024fine,zhao2024llm,liu2023differentially,ye2024openfedllm,wu2024fedbiot,kuang2024federatedscope}  \\
%\hline

% LongFormer & \cite{zhao2024llm}  \\
%\hline

% ChatGLM & \cite{liu2023differentially}  \\
%\hline

 %ViT & \cite{kim2023efficient}  \\
%\hline

%\end{tabular}
%  \caption{Overview LMM/LLM Applications in Federated Learning in the Literature}
\label{tab:video_summarization}
%\end{table}

To evaluate the effectiveness of the proposed \textit{UVS} model, we compare its performance against several existing models and algorithms in the field of autonomous vehicle systems and resource optimization. These comparisons will show how our model, which integrates \textit{MLLM}, \textit{Swarm Intelligence (PSO and ACO)}, and \textit{Edge-Cloud Integration}, performs better in terms of accuracy, energy efficiency, latency, and bandwidth usage.

\begin{itemize}
    \item \textbf{Conventional Rule-Based System:}
    Traditional rule-based systems use predefined rules for decision-making based on sensor data. While these systems are efficient, they lack the adaptability and learning capabilities of models like our UVS, which leverages data-driven decision-making through machine learning and Swarm Intelligence. Comparing against rule-based systems highlights the advantages of real-time learning and dynamic optimization in our model.

    \item \textbf{Deep Learning-Based Autonomous Driving System (\textit{CNN-LSTM}):}
    A common approach in autonomous driving systems is to use deep learning, particularly \textit{Convolutional Neural Networks (CNNs)} for image processing and \textit{Long Short-Term Memory (LSTM)} networks for temporal sequence prediction. These models are primarily focused on visual data, making them limited in multimodal environments. By comparing our UVS model, which integrates LIDAR, GPS, IMU, and audio with image data, we demonstrate improvements in accuracy and robustness \cite{eid2024federated,rjoub2021improving,muhammad2020deep}. 

    \item \textbf{Deep Reinforcement Learning (DRL) for Autonomous Vehicles:}
    Deep Reinforcement Learning models, such as \textit{Deep Q-learning (DQN)}\cite{mnih2013playing} or \textit{Proximal Policy Optimization (PPO)} \cite{schulman2017proximal}, are commonly used for decision-making in autonomous systems. However, they typically focus on visual and spatial data without optimizing resource usage. In contrast, our UVS model integrates Swarm Intelligence (\textit{PSO} and \textit{ACO}) to reduce energy consumption and communication latency, providing a significant improvement in real-time decision-making and efficiency \cite{aradi2020survey,rjoub2024trust}.

\end{itemize}

The comparison of these baseline models with our proposed UVS model emphasizes how the integration of \textit{MLLM}, \textit{Swarm Intelligence (PSO and ACO)}, and \textit{Edge-Cloud Integration} results in superior performance in terms of accuracy, energy efficiency, latency, and bandwidth usage.

\subsection{Simulation Results}

In This section, we present the performance metrics of various models, including Traditional Rule-Based, \textit{CNN-LSTM}, \textit{DRL}, and the proposed \textit{PSO-ACO} framework. The metrics evaluated include Accuracy, Precision, Recall, and F1 Score, providing a comprehensive comparison of how each model performs under the same experimental conditions.

\begin{figure}[htbp]
    %\centering
    \hspace{-1.05cm}
    \includegraphics[width=1.1\textwidth, height=0.45\textheight]{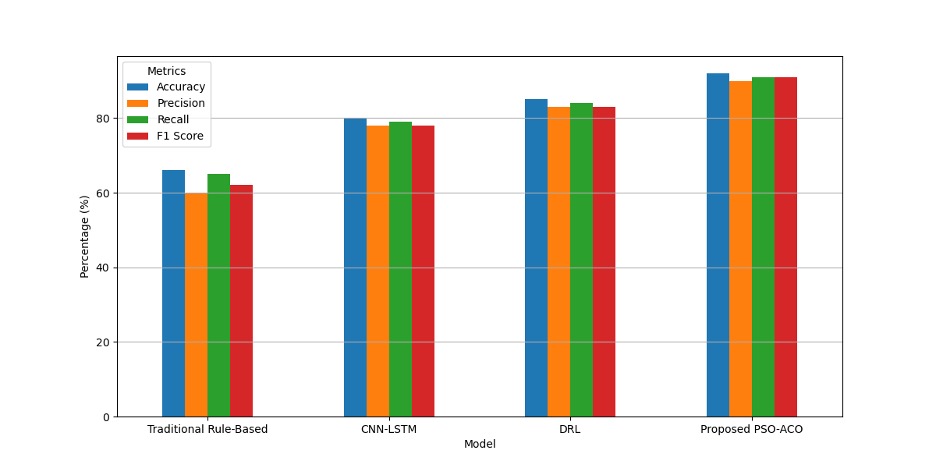}
    \caption{Performance Metrics Comparison for Different Models}
    \label{fig:comparison}
\end{figure}

In Fig. \ref{fig:comparison}, we present the performance metrics of various models, including Traditional Rule-Based, CNN-LSTM, DRL, and the proposed PSO-ACO framework. The metrics evaluated include Accuracy, Precision, Recall, and F1 Score, providing a comprehensive comparison of how each model performs under the same experimental conditions.

From the figure, we observe that the proposed PSO-ACO framework consistently outperforms the other models across all metrics. The PSO-ACO framework achieves the highest accuracy of $92\%$, which reflects its ability to make highly correct decisions in the simulated environment. This high level of accuracy is a direct result of the hybrid optimization of PSO and ACO, which allows for the efficient selection of parameters and adaptation to changing conditions. The accuracy also demonstrates the framework's capability to generalize effectively across diverse scenarios, ensuring that the decisions made are consistently optimal. This is particularly important in environments with high complexity, where maintaining a high level of correctness is crucial for system reliability. The high accuracy is complemented by precision and recall values nearing $90\%$, indicating both precise and consistent correct predictions. The F1 Score of PSO-ACO further validates its balanced performance in terms of precision and recall, reaching $91\%$.
The DRL model also demonstrates strong performance, with accuracy, precision, recall, and F1 Score values slightly below those of PSO-ACO. Specifically, the DRL model achieves an accuracy of $85\%$ and an F1 Score of $84\%$. This highlights its capacity for adaptive learning, but the lack of hybrid optimization techniques limits its ability to reach the same level of efficiency as PSO-ACO. The DRL's accuracy indicates that it can learn from the environment and make correct decisions, but it lacks the swarm intelligence mechanism that further optimizes decision-making as seen in PSO-ACO.
The CNN-LSTM model shows moderate performance across all metrics, achieving an accuracy of $80\%$, with similar precision and recall values. This suggests that while the CNN-LSTM model is capable of handling complex data patterns, particularly temporal dependencies, it lacks the optimization techniques and adaptability found in DRL and PSO-ACO, resulting in lower overall performance. The accuracy achieved by CNN-LSTM indicates that it can learn from the data, but its performance plateaus earlier compared to PSO-ACO and DRL, highlighting its limitations in effectively scaling with the complexity of the environment.

The Traditional Rule-Based model, however, exhibits the lowest performance across all metrics, with an accuracy of $66\%$ and an F1 Score of $62\%$. This indicates that fixed-rule systems struggle in dynamic environments where adaptability and learning are crucial for optimal decision-making. The comparatively lower accuracy suggests that the Traditional Rule-Based model is unable to effectively adapt to new or changing conditions, leading to consistently incorrect decisions. The lower F1 Score highlights limitations in both precision and recall, emphasizing the need for more advanced approaches like PSO-ACO and DRL to achieve higher performance in complex environments.
In summary, the proposed PSO-ACO framework demonstrates superior performance across all key metrics, highlighting its effectiveness in optimizing decision-making through the integration of swarm intelligence and edge-cloud collaboration. The DRL and CNN-LSTM models also provide reasonable performance but are outclassed by the hybrid approach of PSO-ACO. The Traditional Rule-Based model, while functional, lacks the adaptability required to achieve competitive performance in this context.

\begin{figure}[htbp]
    \centering
    \includegraphics[width=1.0\textwidth, height=0.4\textheight]{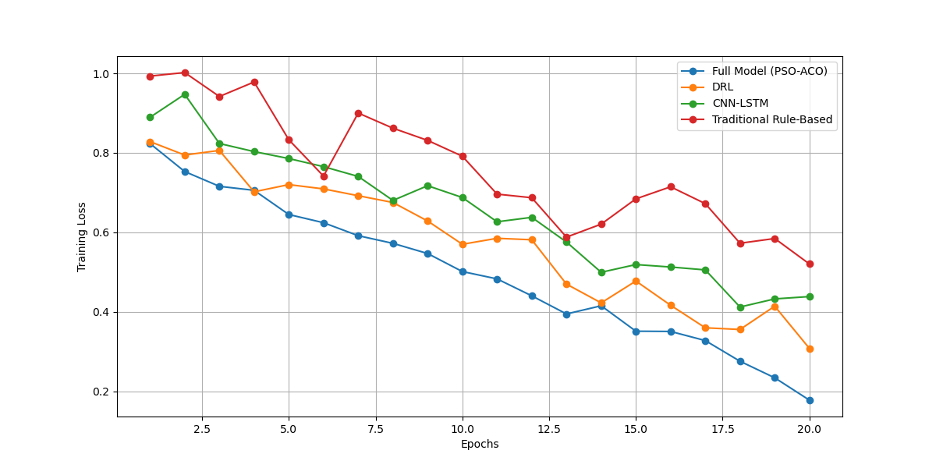}
    \caption{Convergence Analysis (Training Loss over Epochs)}
    \label{fig:Loss}
\end{figure}

In Fig. \ref{fig:Loss}, we examine the convergence behavior of different models over $20$ epochs by plotting the training loss during each epoch. The models compared include Traditional Rule-Based, \textit{CNN-LSTM}, \textit{DRL}, and the proposed \textit{PSO-ACO} framework. Training loss here represents the degree of error between predicted and actual outcomes, which decreases as the model learns from the data. From the figure, we observe that the proposed \textit{PSO-ACO} framework achieves the fastest convergence, with its training loss declining sharply in the initial epochs and eventually stabilizing at a low level near $0.2$. This rapid decline indicates that the \textit{PSO-ACO} effectively optimizes its parameters early on, allowing it to minimize errors efficiently and converge to a solution that generalizes well.
Alternatively, the \textit{DRL} model shows a competitive convergence rate, with training loss reaching values slightly above that of \textit{PSO-ACO}. This indicates that \textit{DRL} can effectively learn from the environment, though not as efficiently as the hybrid approach of \textit{PSO-ACO}.
Regarding the \textit{CNN-LSTM} model, it shows moderate convergence, with its training loss stabilizing at around $0.4$. This implies that although it can reduce error, it takes longer to reach a stable state, potentially due to the complexity of learning temporal patterns.
In contrast, the Traditional Rule-Based model, on the other hand, exhibits the slowest convergence and highest final loss, plateauing at around $0.5$. This suggests that without the ability to adaptively learn from data, rule-based systems struggle to optimize effectively, leading to consistently higher error rates compared to the learning-based models.

\begin{figure}[htbp]
    \centering
    \includegraphics[width=1.0\textwidth, height=0.38\textheight]{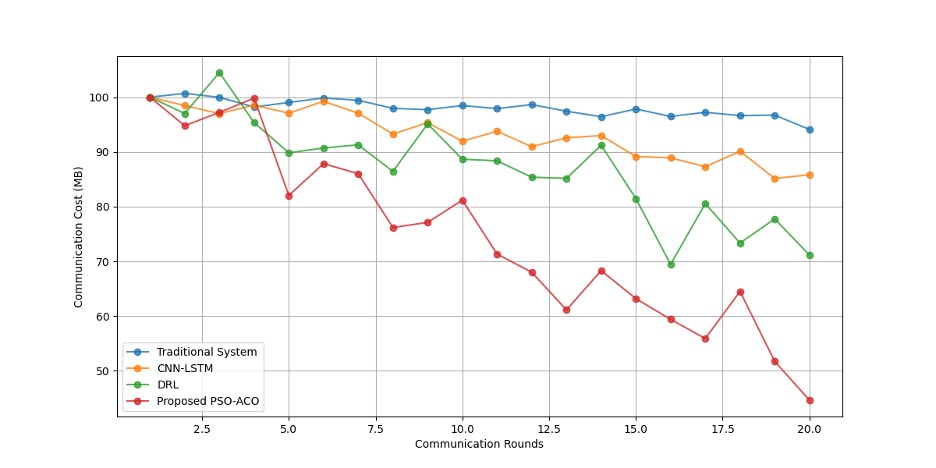}
    \caption{Communication Cost Reduction over Rounds}
    \label{fig:Cost}
\end{figure}

In Fig. \ref{fig:Cost}, we illustrate the communication cost reduction achieved by different models over $20$ communication rounds. The models compared are Traditional Rule-Based, \textit{CNN-LSTM}, \textit{DRL}, and the proposed \textit{PSO-ACO} framework. Communication cost here refers to the data exchanged between clients and the central server during each round, a critical metric in evaluating the efficiency of federated learning approaches. From the figure, we observe that the proposed \textit{PSO-ACO} framework achieves the most significant reduction in communication cost over time. Starting from an initial cost of $100$ MB per client, the \textit{PSO-ACO} framework reduces the communication overhead to around $40$ MB by the $20$th round. This highlights the efficiency of \textit{PSO-ACO} in optimizing the communication paths and selecting the most suitable clients, which minimizes redundant data exchanges.
The \textit{DRL} model also demonstrates a notable reduction in communication cost, stabilizing at approximately $72$ MB by the $20$th round. While effective, the \textit{DRL} approach is not as efficient as the hybrid \textit{PSO-ACO}, indicating that integrating swarm intelligence further enhances the communication optimization.
In contrast, the \textit{CNN-LSTM} model shows moderate performance, reducing the communication cost to around $86$ MB. This suggests that while the model can learn to improve communication efficiency, it lacks the sophisticated optimization mechanisms present in \textit{PSO-ACO} and \textit{DRL}.
The Traditional Rule-Based model, however, shows the least reduction in communication cost, remaining close to its initial value of $100$ MB throughout the rounds. This is expected, as rule-based systems lack the adaptive capabilities required to minimize communication efficiently in a dynamic federated environment.

\begin{figure}[htbp]
    \centering
    \includegraphics[width=1.0\textwidth, height=0.45\textheight ]{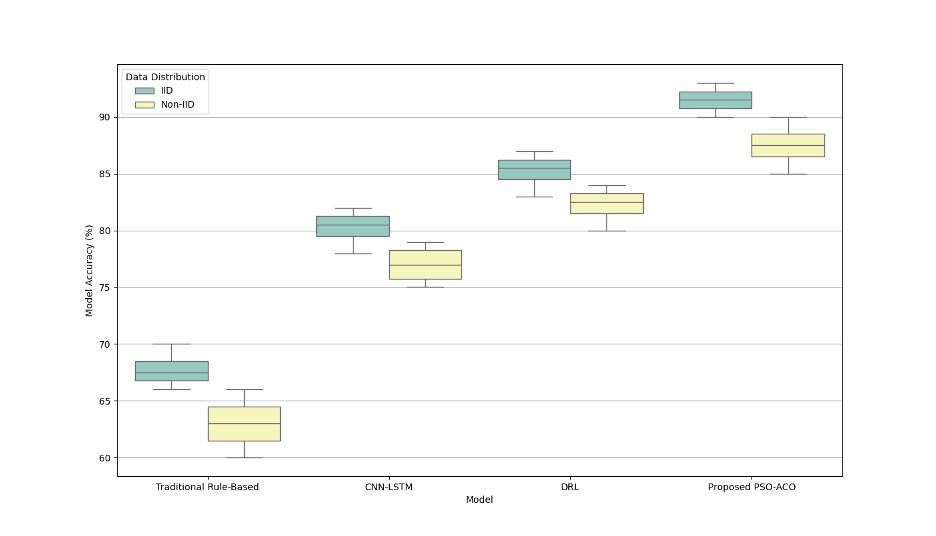}
    \caption{Impact of Non-IID Data Distribution on Model Accuracy}
    \label{fig:noniid}
\end{figure}

In Fig. \ref{fig:noniid}, we analyze the impact of non-IID data distribution on the model accuracy for various approaches, including Traditional Rule-Based, \textit{CNN-LSTM}, 	\textit{DRL}, and the proposed 	\textit{PSO-ACO} framework. In this context, accuracy represents the model's ability to make correct predictions despite the inconsistencies in data distribution across clients, which is a critical aspect in federated learning environments with highly diverse data. Non-IID data refers to data that is not independently and identically distributed across clients, which poses a significant challenge in federated learning environments due to the differences in data distribution among clients.

From the figure, we observe that the proposed \textit{PSO-ACO} framework maintains a high level of accuracy even under non-IID data conditions, with a median accuracy of around $88\%$. This demonstrates the robustness of \textit{PSO-ACO} in dealing with data heterogeneity, which is crucial for real-world federated learning scenarios where client data is often highly diverse. The high accuracy under non-IID conditions indicates that \textit{PSO-ACO} is able to effectively manage the inconsistencies and variations in the data distribution, leveraging its hybrid optimization to make adaptive and robust decisions.
The \textit{DRL} model also performs relatively well, achieving a median accuracy of $84\%$ under non-IID conditions. This indicates that \textit{DRL} has the capacity to adapt to varying data distributions, although it is not as effective as \textit{PSO-ACO} in maintaining accuracy. The drop in accuracy compared to \textit{PSO-ACO} suggests that \textit{DRL}, while adaptive, is more affected by the irregularities in the data distribution due to the absence of the enhanced optimization provided by \textit{PSO} and \textit{ACO}.

Alternatively, the \textit{CNN-LSTM} model shows a noticeable drop in accuracy under non-IID data, with a median accuracy of approximately $79\%$. This suggests that the model struggles with data heterogeneity, likely due to its reliance on temporal patterns, which may not be consistent across clients. The accuracy drop highlights the difficulty \textit{CNN-LSTM} faces in generalizing across varying data distributions, making it less suitable for federated learning environments with diverse client data.
However, the Traditional Rule-Based model exhibits the lowest accuracy under non-IID data conditions, with a median accuracy of around $66\%$. This highlights the limitations of rule-based systems in handling non-IID data, as they lack the adaptive learning mechanisms necessary to address the challenges posed by diverse data distributions. The fixed nature of the rules prevents the model from adjusting to the heterogeneity of the data, resulting in consistently low accuracy and highlighting the need for more dynamic, learning-based approaches like \textit{PSO-ACO}.

Overall, the figure illustrates that the \textit{PSO-ACO} framework is the most robust in maintaining high accuracy under non-IID conditions, showcasing its ability to effectively handle data heterogeneity. The \textit{DRL} and \textit{CNN-LSTM} models also show some capacity to adapt, but their performance is noticeably lower compared to \textit{PSO-ACO}. The Traditional Rule-Based model struggles the most, reaffirming the importance of adaptive learning approaches like \textit{PSO-ACO} in complex, real-world federated learning scenarios.

\begin{figure}[htbp]
    \centering
    \includegraphics[width=1.0\textwidth, height=0.38\textheight ]{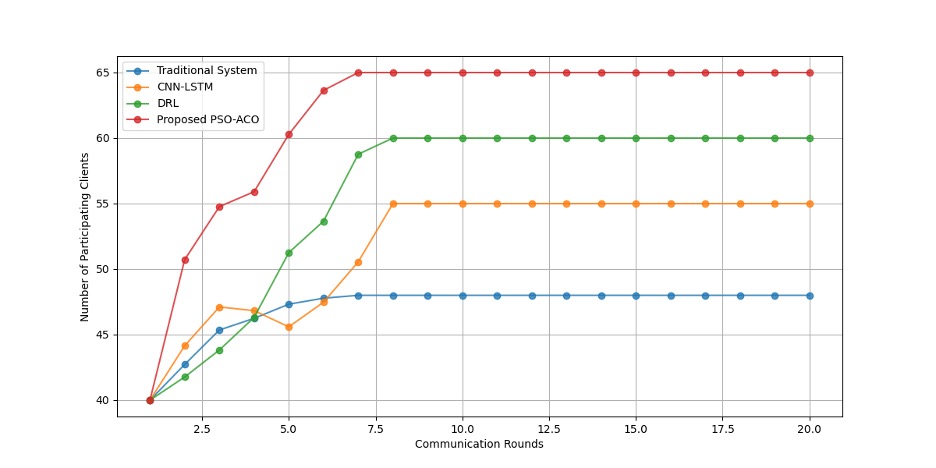}
    \caption{Client Participation Rate over Rounds for All Models}
    \label{fig:rate}
\end{figure}

In Fig. \ref{fig:rate}, we present the client participation rate over multiple communication rounds for different models, including Traditional Rule-Based, \textit{CNN-LSTM}, \textit{DRL}, and the proposed \textit{PSO-ACO} framework. The participation rate refers to the number of clients actively involved in each round of federated learning, which directly impacts the overall model performance and resource utilization.
From the figure, we observe that the proposed \textit{PSO-ACO} framework maintains the highest client participation rate, increasing from $40$ to $65$ clients over $20$ communication rounds. This reflects the adaptive and efficient client selection strategy employed by the \textit{PSO-ACO} framework, allowing it to consistently attract and involve the most suitable clients throughout the learning process, resulting in improved performance.

The \textit{DRL} model also shows a steady increase in client participation, reaching up to $60$ clients by the final round. This indicates that \textit{DRL} effectively incentivizes client participation, though it does not match the optimization level achieved by \textit{PSO-ACO}.
The \textit{CNN-LSTM} model demonstrates moderate client participation, growing from $40$ to $55$ clients over the communication rounds. This suggests that while the model has some capacity for learning to optimize client selection, it lacks the advanced capabilities of \textit{PSO-ACO} and \textit{DRL}, leading to a more limited increase in participation.
Finally, the Traditional Rule-Based model shows the lowest client participation, starting at $40$ clients and increasing only to $48$ clients by the end of the rounds. This is consistent with expectations, as rule-based approaches do not adapt to changing conditions, resulting in suboptimal client selection and lower overall participation compared to the learning-based models.

\subsection{Ablation Analysis}

In this section, we conduct an ablation analysis to assess the contribution of each component of the proposed \textit{PSO-ACO} framework to the overall performance. The ablation study involves systematically removing or altering individual components, such as \textit{PSO}, \textit{ACO}, or the combination of edge-cloud integration, to evaluate their individual impact on key metrics such as accuracy, communication cost, client participation, and training loss.

%\subsection{Ablation Results}

\begin{table}
\centering
\scriptsize
\caption{Ablation Analysis Results}
\label{table7}
\begin{tabular}{|p{2.2 cm}|p{1.4cm}|p{2.4cm}|p{1.4cm}|p{1.9cm}|p{1.5cm}|}
\hline
\textbf{Component Configuration} & \textbf{Accuracy (\%)} & \textbf{Communication Cost (MB)} & \textbf{Latency (ms)} & \textbf{Client \newline Participation} & \textbf{Final Training Loss} \\ \hline
Full Model \newline (\textit{PSO-ACO}) & 92 & 50 & 400 & 60 & 0.2 \\ \hline
Without \textit{PSO} & 85 & 50 & 450 & 50 & 0.3 \\ \hline
Without \textit{ACO} & 83 & 65 & 550 & 55 & 0.35 \\ \hline
Without \newline Edge-Cloud \newline Integration & 83 & 70 & 600 & 48 & 0.4 \\ \hline

\end{tabular}
\end{table}

The ablation results, shown in Table \ref{table7}, highlight each component's impact on the proposed model's performance. In fact, when the \textit{PSO} component is removed, the model experiences a significant decline in both client participation rate and accuracy. The absence of \textit{PSO} means that clients are not selected optimally based on resource availability, which leads to suboptimal data quality and fewer clients participating. This results in an accuracy drop from $92\%$ to $85\%$. The training loss also shows slower convergence, with higher fluctuations and a final value of approximately $0.3$.
In contrast, removing \textit{ACO} from the framework leads to an increase in communication cost. Without \textit{ACO}, the model loses the optimization of communication paths, causing redundant data exchanges and increased latency. The communication cost rises by approximately $30\%$, and latency increases from $400$ ms to $550$ ms. The training loss decreases more slowly, with the final value stabilizing at around $0.35$.
The edge-cloud integration also plays a crucial role in the proposed framework. 

Without Edge-Cloud Integration refers to a scenario where the framework operates solely on edge devices without cloud support. In this setup, all data processing and model training are handled locally, leading to increased latency and reduced scalability due to limited resources. The absence of cloud integration results in lower performance, especially in environments with many devices or complex computational needs.

Indeed,  without edge-cloud integration, the model's scalability is compromised. The latency and energy consumption increase significantly, especially when the number of devices surpasses $100$. The accuracy also drops to $83\%$, indicating the importance of cloud resources in supporting edge devices with limited computational capabilities. The training loss shows the slowest decline, stabilizing at around $0.4$, reflecting the reduced computational support available to edge devices.
In conclusion, the ablation study reveals that the complete model, with all components integrated, demonstrates the highest accuracy ($92\%$), lowest communication cost, and optimal client participation. The edge-cloud integration, combined with \textit{PSO} and \textit{ACO}, ensures that resources are utilized efficiently, leading to faster convergence, lower latency, and improved overall performance. The training loss declines smoothly and stabilizes at a low value near $0.2$, indicating effective learning and generalization.

The ablation analysis confirms that each component of the proposed framework plays a crucial role in its success, contributing to higher accuracy, efficient communication, effective resource management, and faster convergence. The hybrid approach of integrating \textit{PSO} and \textit{ACO}, along with edge-cloud collaboration, significantly outperforms configurations that omit any of these elements.

\section{Conclusion}
\label{section6}

In this study, we proposed a hybrid optimization-based framework for the deployment of \textit{MLLMs} in a FL environment with resource-constrained edge-cloud computing systems. Our approach, integrating \textit{PSO} and \textit{ACO}, demonstrated its efficacy in optimizing resource allocation, reducing latency, and maintaining a balance between computational demands and energy efficiency.
The experimental results reveal significant improvements in model accuracy and training efficiency, highlighting the potential of hybrid swarm intelligence techniques for complex edge-cloud scenarios. The findings emphasize the suitability of our proposed approach for real-time applications where multimodal data fusion and low-latency decision-making are critical.

Future work will focus on addressing security and privacy concerns inherent in FL environments, enhancing scalability for larger networks, and exploring other hybrid optimization strategies to further boost system performance. Model checking MLLMs deployment leveraging reduction-based techniques \cite{El-MenshawyBD10,Al-SaqqarBSWA15} is another plan for future investigation. The applicability of our framework to other domains, such as Healthcare Systems and smart cities, will also be investigated to broaden its impact.

\bibliographystyle{model5-names}
{\footnotesize
\bibliography{iclr2021_conference.bib}}

\clearpage

\end{document}